\documentclass[sigconf,nonacm]{acmart}
\pdfoutput=1
\usepackage{algorithm}
\usepackage{algorithmic}
\usepackage{booktabs} 
\usepackage{multirow}
\usepackage{amsmath}
\usepackage{amssymb}
\AtBeginDocument{%
  }

\begin{document}

\title{Perseus: Leveraging Common Data Patterns with Curriculum Learning for More Robust Graph Neural Networks}

\author{Kaiwen Xia}
\email{xiakaiwen@nudt.edu.cn}
\orcid{0009-0004-2367-3088}
\affiliation{%
  \institution{National University of Defense Technology}
  \city{Changsha}
  \country{China}
}

\author{Huijun Wu}
\email{wuhuijun@nudt.edu.cn}
\affiliation{%
  \institution{National University of Defense Technology}
  \city{Changsha}
  \country{China}}

\author{Duanyu Li}
\email{liduanyu19@nudt.edu.cn}
\affiliation{%
  \institution{National University of Defense Technology}
  \city{Changsha}
  \country{China}
}

\author{Min Xie}
\email{xiemin@nudt.edu.cn}
\affiliation{%
  \institution{National University of Defense Technology}
  \city{Changsha}
  \country{China}
}

\author{Ruibo Wang}
\email{ruibo@nudt.edu.cn}
\affiliation{%
  \institution{National University of Defense Technology}
  \city{Changsha}
  \country{China}
}

\author{Wenzhe Zhang}
\email{zhangwenzhe@nudt.edu.cn}
\affiliation{%
  \institution{National University of Defense Technology}
  \city{Changsha}
  \country{China}
}

\begin{abstract}
Graph Neural Networks (GNNs) excel at handling graph data but remain vulnerable to adversarial attacks. Existing defense methods typically rely on assumptions like graph sparsity and homophily to either preprocess the graph or guide structure learning. However, preprocessing methods often struggle to accurately distinguish between normal edges and adversarial perturbations, leading to suboptimal results due to the loss of valuable edge information. Robust graph neural network models train directly on graph data affected by adversarial perturbations, without preprocessing. This can cause the model to get stuck in poor local optima, negatively affecting its performance. To address these challenges, we propose Perseus, a novel adversarial defense method based on curriculum learning. Perseus assesses edge difficulty using global homophily and applies a curriculum learning strategy to adjust the learning order, guiding the model to learn the full graph structure while adaptively focusing on common data patterns. This approach mitigates the impact of adversarial perturbations. Experiments show that models trained with Perseus achieve superior performance and are significantly more robust to adversarial attacks.
\end{abstract}

\keywords{Graph neural networks, adversarial defense algorithms, robustness analysis, curriculum learning}

\maketitle

\section{Introduction}
Network structures, including social networks, citation networks, and molecular networks, are pervasive in the real world and serve as integral components in numerous fields. Graph neural networks (GNNs) represent an advanced tool for processing graph data, and have demonstrated excellent performance in several tasks. These include link prediction, node classification\cite{ICL_SSL, MGCN}, and graph classification. However, despite the considerable advances made by GNNs, recent research has demonstrated their susceptibility to adversarial attacks. Even minor and imperceptible alterations to graph structures have the potential to result in a notable decline in the performance of GNNs in real-world applications, which presents a significant challenge to the reliability and security of GNNs.

To address this issue, recent studies have approached it from different perspectives. On the one hand, they attempt to use preprocessing-based methods to identify and remove malicious perturbations in graph data. On the other hand, they strive to construct more robust GNN models. 

Preprocessing methods \cite{chen2020iterative, entezari2020all, jin2020graph, zheng2020robust, luo2021learning} typically rely on inherent structural properties of graphs, assuming "sparsity" or "low-rankness." Based on these assumptions, recent work\cite{dai2018adversarial, zhang2019comparing, zugner2018adversarial, entezari2020all, ioannidis2019graphsac} commonly employ deterministic techniques to optimize the graph structure, such as processing through Singular Value Decomposition (SVD), or dynamically removing "redundant" edges based on the performance of downstream tasks on specific sparsified structures\cite{chen2020iterative, jin2020graph, luo2021learning}. However, such modifications to the graph topology may result in the loss of potentially important information contained in the removed edges. Moreover, due to limited capability in distinguishing between origin and perturbed edges, this approach may lead to the erroneous removal of valuable edge data. More critically, modifications to the graph structure are often optimized for optimal performance on the training set, which can easily lead to overfitting issues.

Robust graph neural networks enhance resistance to adversarial attacks through adversarial training or modifications to the model architecture. Adversarial training, as cited in \cite{jin2020adversarial, li2020deeprobust, deng2023batch, chen2019can, ennadir2024simple}, increases model robustness by incorporating adversarial samples into the training data. Additionally, a portion of the research focuses on innovations in model architecture \cite{zhu2019robust, tang2020transferring, jin2020graph, zhao2024adversarial}, aiming to design network structures that are more resilient to perturbations. Although these strategies have bolstered the model's robustness to some degree, the existence of adversarial perturbations may still cause the model to be easily stuck in poor local optimum during the training process.
Against this backdrop, we seek inspiration from cognitive science. Research in cognitive science indicates \cite{elman1993learning, paszke2019pytorch} that humans achieve deeper understanding and better generalization capabilities by learning from simple concepts to complex ones. Drawing on this concept, the method of Curriculum Learning (CL) \cite{bengio09} emerged. This method, by strategically sequencing the difficulty of learning tasks, is able to smooth the optimization process and mitigate the instability of gradients, effectively preventing the model from being trapped in local optima, and guiding it towards a progressively superior solution space. Multiple studies\cite{bengio09, gong2015curriculum, han2018co, jiang2018mentornet, shrivastava2016training, weinshall2018curriculum} have shown that an effective sequence of data sample learning can significantly enhance a model's generalization and robustness. 
In previous research, the focus has primarily been on deep neural networks, with little to no work addressing the incorporation of curriculum learning strategies for Graph Neural Networks (GNNs). Recent studies indicate that GNN models are vulnerable to adversarial attacks on graph structures. Consequently, the effective introduction of curriculum learning (CL) strategies can guide GNN models to learn more accurate and robust graph structures, thereby enhancing the model's performance and security.
Hence, we introduce the curriculum learning paradigm into the graph defense model and innovatively propose a novel curriculum learning based graph neural network framework named Perseus. We have designed a dynamic and adaptive mechanism that not only filters out reliable samples for efficient training but also adjusts the rate and timing of learning new samples in real-time. This mechanism enables the model to progress from simple to complex, gradually learning the correct graph structure data, significantly enhancing the model's generalization capabilities and adversarial robustness.

The design philosophy of Perseus is to first concentrate on learning the edges that are prevalent and undisturbed edges, enabling the model to develop an accurate understanding of the common graph structures. Building upon this foundation, the framework progressively extends to handling more complex edges. Since perturbation information is not predominant in the graph, this approach ensures that the model can establish a solid foundation based on common data patterns during the early learning phase. Moreover, through the design of an adaptive learning pace, it effectively prevents the model from being trapped in locally optimal solutions guided by a minority of attack information, thus guiding the model towards a robust solution space.

Our contributions are as follows: Firstly, we introduce curriculum learning into graph neural networks for the first time, constructing an adaptive graph structure learning framework based on curriculum learning, which can be integrated with various graph defense algorithms. Secondly, we propose a global metric to quantify the difficulty of edges and evaluating the performance of various metric. Thirdly, we design an adaptive edge selection strategy that dynamically adjusts based on the training state. Finally, through extensive experiments on five real-world datasets, we have verified the superior performance of Perseus under adversarial attacks, particularly with high perturbation ratios, demonstrating the effectiveness of the curriculum learning strategy in enhancing the robustness of GNNs.

\section{Related Work}
In this section, we will briefly review two relevant research directions.
\subsection{Curriculum Learning (CL)}
The concept of curriculum learning (CL) in the machine learning field was first introduced in 2009 \cite{bengio09}, aiming to improve the performance of the model by progressively adding easy to difficult samples during the training of the model. Many works \cite{jiang14, jiang15, zhou2020robust} propose supervised measures to determine curriculum difficulty, such as the diversity of samples \cite{jiang14} or the consistency of model predictions \cite{zhou2020robust}. Meanwhile, many empirical and theoretical studies, from different perspectives, have explained why CL improves the generalisation ability of models \cite{han2018co, jiang2018mentornet}. There are also theoretical explanations for the denoising mechanism of CL learning in noisy samples \cite{gong2015curriculum}. Although CL has achieved good performance on many tasks, most of the existing designs for CL strategies target independent data, such as images, and very little work has used CL strategies for samples with dependencies. Existing works \cite{ju2024comprehensive, li2023curriculum, li2024graph} simply consider nodes as independent samples to apply CL strategies, ignoring the graph structure information, and cannot handle the correlation between data samples well. Moreover, most of these models are based on heuristic-based sample selection strategies \cite{chu2021cuco, li2024graph, wei2023clnode}, which severely limits their generalisability.

\subsection{Graph Adversarial Defense}
To improve the reliability and safety of GNNs, current research on improving model robustness falls into two main categories. One category focuses on the preprocessing stage of the graph  \cite{Said10, entezari2020all}, i.e., the processing of the graph such as cropping of perturbed edges, etc., before input to the model. Another category of research focuses on the model itself, hoping to train more powerful GNNs through adversarial or constraints \cite{jin2020adversarial, li2020deeprobust, deng2023batch, chen2019can, ennadir2024simple} or to make modifications to the GNN architecture \cite{zhu2019robust, tang2020transferring, zhao2024adversarial}.

For preprocessing methods, Xu et al. \cite{xu2018characterizing} proposed a method based on graph generative models, utilizing link prediction as a preprocessing step to further detect potential malicious edges. On attributed graphs, based on the observation that attackers prefer to add edges rather than remove them, and that edges are often added between dissimilar nodes, Edog \cite{xu2023edog} samples subgraphs from the poisoned training data and then uses outlier detection methods to detect and filter adversarial edges. GCN-Jaccard \cite{wu2019adversarial} preprocesses the data by calculating the Jaccard similarity between adjacent nodes in the spatial domain to eliminate the impact of adversarial perturbations. On the other hand, numerous works have considered modifying the model structure to propose robust graph neural networks for defending against graph adversarial attacks. Some studies \cite{jin2020adversarial, li2020deeprobust} are based on two opposing objective functions (for minimization and maximization, respectively) and gradually enhance model performance through a continuous minimax iteration process. NoisyGNN \cite{ennadir2024simple} counters attacks by dynamically introducing stochastic noise into graph neural networks, exhibiting model-agnosticism and a favorable balance of performance.
The Batch Virtual Adversarial Training (BVAT) algorithm \cite{deng2023batch} aims to gain insights into the connection patterns between nodes in the graph by generating virtual adversarial perturbations, thereby enhancing the smoothness of the node classifier's output distribution. A study \cite{zhao2024adversarial} proposes a graph neural network architecture based on conservative Hamiltonian neural flows, which significantly enhances the model's robustness against adversarial attacks by incorporating Lyapunov stability principles.

Although existing defense algorithms have undergone extensive design and improvement on graph structures or model architectures, providing a certain level of protection against low-intensity attacks, several core issues remain. 

Firstly, the reconstruction process of the graph structure during the application of preprocessing methods inevitably leads to a significant loss of valuable information, which in turn undermines the performance of the algorithm to a certain extent.

Secondly, in the training of robust graph neural networks, the perturbation introduced by attack data may cause the model to learn based on incorrect perceptions, making it prone to converging to erroneous local optima. This issue is particularly pronounced when the proportion of attack data is high, as the learning process of the model becomes more susceptible to the influence of attack data, thereby reducing the overall robustness and accuracy of the model.

In light of this, there is an urgent need to develop a novel defensive methodology that preserves the integrity of graph data information while effectively circumventing the potential misleading impact of attack data on model predictions.

\begin{figure}[t]
\centering
\includegraphics[width=\columnwidth]{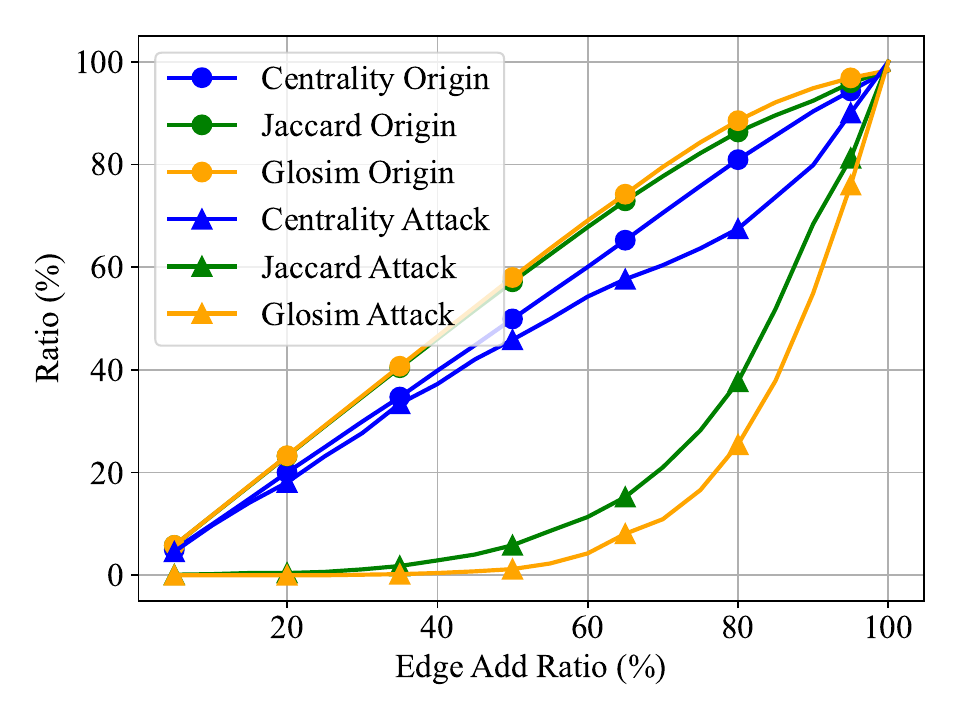} 
\caption{Performance of different metrics on Cora under Metattack with 20\% perturbation rates}
\label{observationexperiment}
\end{figure}
\section{Notation}

A graph is defined as $\mathcal G = (\mathcal V, \mathcal E)$, where $V = \{v_1, \cdots, v_n\}$ denotes the set of nodes and $\mathcal E \in \mathcal V \times \mathcal V$ denotes the set of edges. The edges connecting nodes $v_i$ and $v_j$ are represented by $e_{i,j}$. The number of nodes and the number of edges are denoted by $|\mathcal V| = n$  and $|\mathcal E| = m$, respectively. Let $X \in \mathbb R^{n \times d_f}$ be the feature matrix, where the $i$-th row of $x_i$ represents the $d_f$-dimensional feature vector of node $v_i$. The label $y_v \in C$, which belongs to the set of class C, denotes the class to which the node belongs. The adjacency matrix of $\mathcal G$ is denoted as $A \in R^{n \times n}$, which is an $n \times n$ matrix where $A_{ij} = 1$ if $e_{i,j} \in \mathcal E$, and $A_{ij} = 0$ otherwise. Using the aforementioned feature and adjacency matrices, the graph can also be written as $\mathcal G = (A, X)$.

\begin{figure*}[t]
  \centering
  \includegraphics[width=0.9 \textwidth]{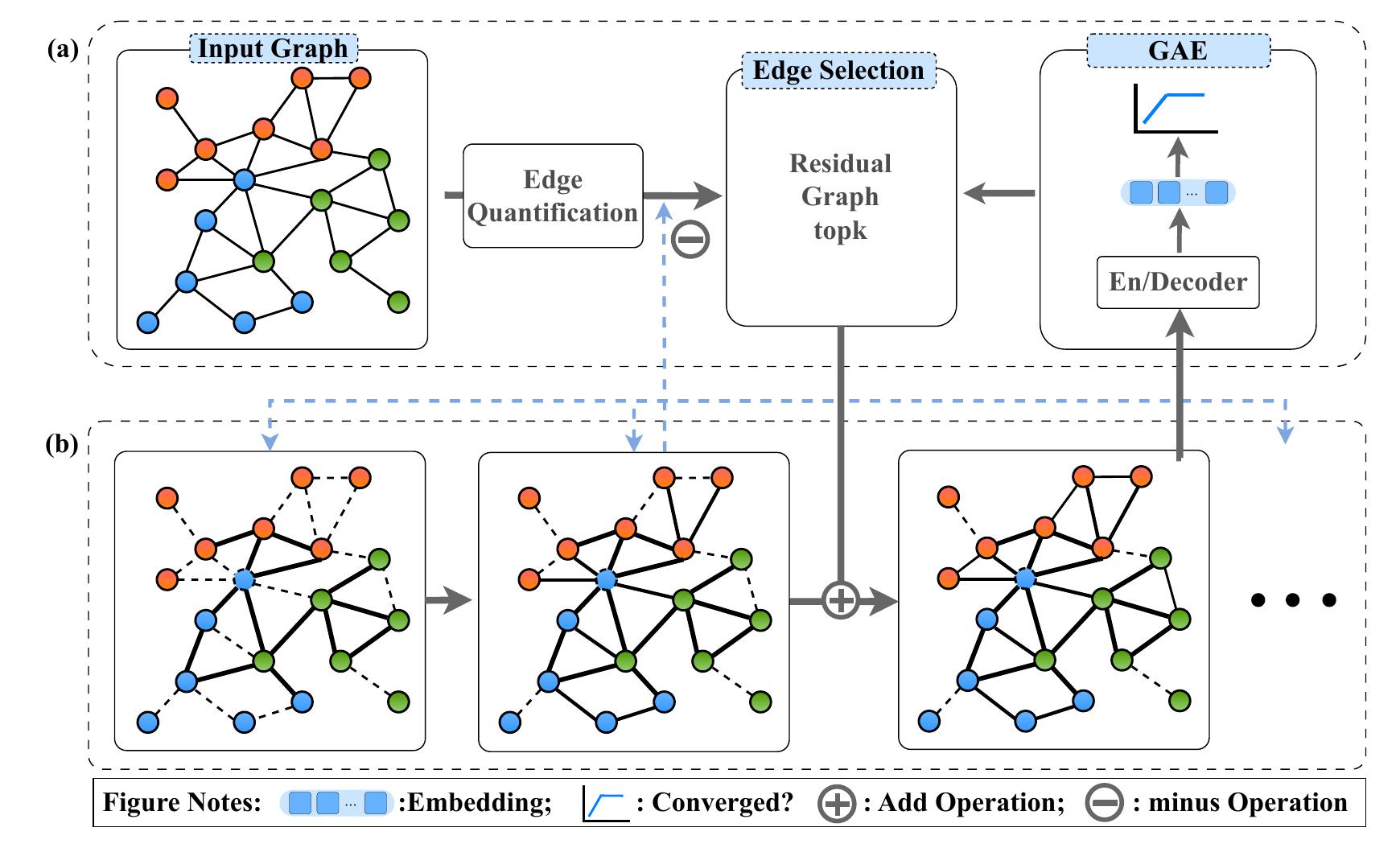} 
  \caption{The overall pipeline of Perseus. (a) Quantifying the edge difficulty based on the original image, the edge selection module takes topk edges based on the current residual graph. Furthermore, the residual graph is obtained from the edge quantization index and the current selected edge. The timing of performing edge selection is subsequently determined by judging the model convergence state. (b) Iterative learning process of Perseus. The model is trained starting from the initial simple structure and gradually adds more multiple optimal edges until the training structure converges to the input structure.}
  \label{Perseus}
\end{figure*}

\section{Methodology}
This section details our proposed method, called Perseus. As shown in Figure~\ref{Perseus}, Perseus consists of two main components: the Adaptive Edge Selection (AES) module and the Graph Auto-Encoder Representation Learning (GAE) module, which interact dynamically during training. Within the edge selection module, the timing for adding edges is dynamically determined by the training status of GAE. Subsequently, the edge data selected by AES is integrated into GAE for further learning. These two primary components, through their mutual promotion and learning, increasingly reinforce each other’s capabilities over time.

\subsection{Edge Quantization}
Given that the majority of attacks in graph neural networks are oriented towards edges, it is intuitively believed that the perturbed edges generated by such attacks are more difficult to navigate compared to the clean edges in the original graph. However, the importance of different edges varies in the original graph, which introduces a degree of complexity. 
Therefore, we propose a global metric to measure the importance of edges, termed Global Homophily (GloHom), and have designed an observational experiment to evaluate the different edge quantification metrics. In graph neural networks, there are numerous metrics for quantifying the difficulty of edges:

{\bf PageRank centrality} PageRank centrality is a metric used to quantify the relative importance of nodes within a web graph. It was originally proposed by Larry Page and Sergey Brin, the founders of Google Inc. The significance of each web page is evaluated based on the linking relationships between pages, where by the more significant pages are assigned higher PageRank values. The mathematical formula is as follows:
\begin{equation}    
\label{equ:1}
PR(u) = (1-d) + d(\sum_{v \in N(u)} PR(v)/L(v) )
\end{equation} 

Where $PR(u)$ denotes the PageRank value of node $u$, $d$ is the damping coefficient, which generally takes the value of 0.85 and indicates the probability of a user randomly jumping to other pages. $N(u)$ is the set of neighbouring nodes of $u$, $PR(v)$ denotes the PageRank value of node $v$, and $L(v)$ denotes the out-degree (i.e., the number of links that point to other nodes) of node $v$. With this formula, we can compute the PageRank value of each node in the network graph, which in turn can be used to compute the centrality of the edges connected between the nodes through statistics such as sum().

{\bf Jaccard Similarity} GCN models are highly reliant on the aggregated information of a node's neighbourhood for prediction. Consequently, those seeking to launch an attack will typically add nodes connected to the target node that exhibit markedly disparate features, with the objective of modifying the aggregated features of the neighbourhood and thereby misleading the model. Consequently, (Wu et al. 2019) have put forth a metric to quantify the resemblance between features, utilising the Jaccard coefficient. The Jaccard coefficient is a measure of the similarity between two set, where in the elements within each node's feature vector are regarded as an set in a graph to ascertain the degree of overlap between the features of two nodes:
\begin{equation}    
\label{equ:Jaccard}
\mathcal{J}_{u,v} = \frac {\sum_{i=1}^n u_{i}v_{i}}{\sum_{i=1}^n u_{i}+v_{i}-u_{i}v_{i}}
\end{equation} 
Where $u$ and $v$ are two nodes with n binary features, the Jaccard similarity score measures the overlap that $u$ and $v$ share with their features. Each feature of $u$ and $v$ can either be 0 or 1.

{\bf Global Homophily} The Jaccard similarity serves as a node-level defense for local structures, whereas many attack methods target the graph with global attacks. Additionally, some adaptive attack strategies bypass immediate neighbors to indirectly perturb second-order neighbors. In light of these considerations, we introduces a metrics called global homophily (glohom). In essence, it is based on the Jaccard index to assess the impact of removing a specific edge on the message passing of graph neural networks, and employs gradient descent to identify the perturbed edges. Initially, model the k-order homophily of graph based on the Jaccard index:

\begin{equation}    
\label{equ:Global Homophily} 
\mathrm{Hom}^{(k)}= <\mathbf{A}^k,\mathcal{J}>
\end{equation} 
where $\mathbf{A}$ denotes the adjacency matrix, $\mathcal{J}$ is the Jaccard similarity matrix, and $<\cdot,\cdot>$ representing the product within the matrix. 
The global homophily of a graph is obtained by aggregating the homophily across different order.
\begin{equation}  
\label{equ:2}
\mathrm{Hom}=\sum_{k=0}^{K} \alpha^{k} \mathrm{Hom}^{(k)}
\end{equation}  
where $\alpha \in [0,1]$ is the average retention rate of the residual information after information aggregation.

The utilisation of gradient descent based on global homophily enables the identification and elimination of edges that propagate the most misleading information. In other words, the objective is to identify an edge that maximises the difference in homogeneity. This can be expressed mathematically as follows:
\begin{equation}  
\label{equ:gradient descent of Homophily}
\mathop{\mathrm{argmax}}\limits_{\Delta \mathbf{A}} \Delta \mathrm{Hom}
\end{equation} 

Following a series of mathematical derivations and approximations, the final result can be obtained:
\begin{equation}
\label{equ:Global sim}
\mathop{\mathrm{argmax}}\limits_{\Delta \mathbf{A}} \Delta \mathrm{Hom} \approx \mathop{\mathrm{argmax}}\limits_{\Delta \mathbf{A}} <M \mathcal{J} M, -\Delta \mathbf{A}>  
\end{equation}
Where $M = (I - \alpha\mathbf{A})^{-1}$. The detailed mathematical derivation is presented in the appendix.

As only the ordering of the edges is required in order to streamline the computation, The edges corresponding to the smallest values of $n$ in the matrix $M \mathcal{J} M$ can be identified directly. An ascending ordering of the corresponding values of these edges can be employed as a quantitative measure of the difficulty of the edges.

{\bf Observation experiment} Due to the unreliability of perturbed data in the attacked graph data, we design an observation experiment to observe the recognition ability of different metrics for perturbed edges. Firstly, based on different metrics, we obtained the quantitative value of the edge, that is, the degree of difficulty $d_{ij}$ of the edges. Then, all edge values are sorted in descending order, and a list $[dij] \in Rm$, representing the degrees of difficulty of the edges ordered from easy to difficult, is obtained. The rules for calculating the ratio of perturbed edges are as follows:
\begin{equation}    
\label{equ:1}
r_p= \frac{|\{ topk[d_{ij}] | e_{ij} \in  e_p \}|}{|\{ topk[d_{ij}] \}|}
\end{equation} 
Where $e_{ij}$ is the edge connected to vertices nodes $i$ and $j$, $e_p$ is the set of perturbed edges, $k = r \ast m$, where $r$ is the set of discrete points in the interval [0, 1] at 0.05 intervals, and $m$ is the total number of edges in the attacked graph. That is, for the simplest edges from the different metrics, how many perturbed edges are contained in them, and also the same logic is used to figure out the percentage of clean edges $r_o$. Intuitively, we want the perturbing edges to be "difficult". As shown in Figure~\ref{observationexperiment}, we can see that the GloHom metric is the best at discriminating between clean edges and perturbed edges. It is also more in line with the requirements of this work task, that is, to learn important and clean edges first, to build a robust framework, and then gradually add edges on the basis, and have certain judgement and counteracting ability when encountering perturbed edges.

\subsection{Adaptive Edge Selection}
In curriculum learning, determining the appropriate timing to introduce training samples is a significant challenge, and numerous studies have focused on this issue. This work presents a dynamic adaptive edge selection strategy (AES) based on the immediate training state. This strategy faces three primary questions: when to add edges, which edges to add, and how to add edges?

Specifically, by monitoring the model's performance on the validation set, the strategy triggers an early stopping mechanism once the performance improvement plateaus or begins to decline. This indicates that the model may be approaching or has reached the upper limit of its learning capacity. This dynamic learning termination strategy ensures that the model effectively prevents overfitting to the training data while fully learning the graph structure information. It aids the model in better absorbing and processing more complex graph structure information in subsequent learning phases, thereby enhancing the model's generalization and robustness.

\textbf{Which Edges to Add?} Based on the visual analysis results of observational experiments, it can be observed that as the proportion of edges increases, the slope of the curve representing the proportion of perturbation edges continues to rise. In other words, with the uniform expansion of the edge selection range, the growth rate of the number of perturbation edges gradually increases. Consequently, we have implemented a decremental approach to the model's learning rate, gradually reducing the proportion of new edges added as the training process progresses. This measure is intended to ensure that the number of perturbation edges added in each iteration is kept within a reasonable range, thereby maintaining the model's stable learning capacity. That is to say, as the number of edge additions increases, the proportion of edges added is as follows:
 
\begin{equation}    
\label{equ:1}
r^{(s)} = max\{(r^{(s-1)}; \lambda), 0.05\}  \\
s.t. r^{(s)} \leq r^{(s-1)}
\end{equation} 

Where $r^{(s)}$ is the ratio of edges added at the \textit{s}-th added edge. So the total number of edges incorporated in each iteration is as follows:
\begin{equation}    
\label{equ:1}
k^{(s)} = \sum_{i=1}^s r^{(s)} \ast m
\end{equation}

Existing work on independent data typically quantifies the difficulty level using supervised metrics, such as the loss of training samples, but such metrics are not available for edges. Therefore, in the context of attacked graph sample data, based on observational experiments, this work employs the GloHom metric to measure the difficulty of edges. The results obtained from Equation (5) are normalized and then sorted in descending order:
\begin{equation}    
\label{equ:1}
[g_{ij}] \in R^m
\end{equation}

Where $g_{ij}$ represents the normalized GloHom metric. Based on the quantification results, new edges are incorporated, thereby updating the graph structure that the model needs to fit: 
\begin{equation}    
\label{equ:z}
A^{(s)} = topk^{(s)}[h_{ij}]
\end{equation} 

\textbf{How to Add Edges?} Intuitively, as the edge selection model iterates, the edges selected later are relatively less important or perturbed edges. Consequently, this study also designs a mechanism for edge weight decay, such that as more challenging edges are added to the model, their weights are gradually reduced. This strategy effectively limits the negative impact of perturbed edges on the model's learning process, ensuring that the model can maintain stable learning performance and generalization ability while absorbing new information. The update method for the edge weight $ew^{(s)}$ is as follows:
\begin{equation}    
\label{equ:z}
ew^{(s)} = (\sum_{i=1}^s A^{(s)}) / s
\end{equation}

\begin{algorithm}[tb]
\caption{Perseus}
\label{alg:algorithm}
\textbf{Input}: Node features $X$, adjacency matrix $A$\\
\textbf{Parameter}: decay $\lambda$, warmstart $r^{(0)}$\\
\textbf{Output}: Parameters $w$ of GAE f
\begin{algorithmic}[1] 
\STATE Initialize $w^{(0)}$, $A^{(0)}$
\IF {$A^{(s)} \neq A$}
\STATE $w^{(t)} = argmin_w L(f(X, A^{(s-1)}; w^{(t-1)}), y)$
\IF {Converged}
\STATE $r^{(s)} = (r^{(s-1)}; \lambda)$
\STATE $A^{(s)} = EdgeSelection(A^{(s-1)}, r^{(s)})$
\STATE $A^{(s)} = (\sum_{i=1}^s A^{(s)}) / s$
\ENDIF
\ENDIF
\STATE \textbf{return} solution
\end{algorithmic}
\end{algorithm}

\subsection{Model Training}
To cooperate with the GAE model and the edge selection module, we trained and performed both components iteratively (see Algorithm 1). The algorithm uses the node feature matrix $X$, the original adjacency matrix $A$ as input, and the hyperparameter $\lambda$ to control the decay of the addition rate. The initialization of $A^{(0)}$ is obtained by the warm edge proportion and the edge quantized value. After $w$ and $A$ are initialized, the GAE and edge selection modules are performed alternately. After the current model converges, the ratio of adding edges (step 5) is calculated, the adding operation (step 6) is then performed, and the edge weights are updated for the following model training (step 7). This iterates until all the edges are added.

\section{Experiment}
\subsection{Setup}
In this section, we evaluate the performance of Perseus against non-target adversarial attacks on graphs. Experiments were performed on a server equipped with the Nvidia ™ (NVIDIA®) Tesla V100-SXM2 32GB GPU. Experimentally, we evaluated the performance of Perseus compared to state-of-the-art defense methods.
\subsubsection{Datasets}

The experiments were conducted on five commonly used benchmark datasets, namely Cora, Citeseer, Photo, Computers and Pubmed. The specifics of the dataset are presented in Table~\ref{DatasetSummary}. In the curriculum of our experiments, we have elected to take the maximum connected subgraph for each of the datasets under consideration. The datasets are randomly divided into three subsets: 10\% is used for training, 10\% for validation, and the remaining 80\% is used for testing. This division provides a rigorous environment for evaluating model performance.
\subsubsection{Generating Adversarial Attacks}

In order to ascertain the performance of Perseus, an attack test was first conducted on graph data. In order to test the efficacy of the aforementioned methodology, two global attack strategies were employed: the meta-attack \cite{Metattack} and the projected gradient descent (PGD) attack \cite{PGD}. In order to implement the meta-attack, a simplified two-layer GCN was employed as an agent model to direct the attack. Furthermore, the agent model comprises a 64-node hidden layer, and the model's generalisation ability is enhanced by setting a dropout rate of 0.5, thereby reducing the risk of overfitting. In addition, the learning rate was set to 0.01, the weight decay was set to 0, and an early stopping strategy was employed to train for 1000 epochs in order to prevent overfitting. In the meta-attack, the self-trained meta-gradient method was employed, with the momentum parameter set to 0.9. In order to test the efficacy of the PGD attack, we implemented the two aforementioned adversarial attack methods on the five datasets, namely Cora, Citeseer, Photo, Computers and Pubmed, on multiple occasions. During the attack, the perturbation ratio of the edges in the graph was adjusted in a gradual manner, increasing from 5\% to 25\% in steps of 5\%. Following the completion of the attack, the graph data was tested using GCN with the same configuration as the meta-attack agent model. The results of the attack experiments are presented in detail in Table~\ref{AttackResult}, and these attacked graph data will be used as a benchmark for subsequent defence evaluations.
\begin{table}[t]
  \centering  
  \caption{Dataset Summary}  
  \label{DatasetSummary}  
  \fontsize{9pt}{11pt}\selectfont
  \begin{tabular}{lrrrr}  
  \toprule  
  \textbf{Dataset} & \textbf{Nodes} & \textbf{Edges} & \textbf{Classes} & \textbf{Features} \\ \midrule  
Cora             & 2,708          & 5,429          & 7                & 1,433               \\  
Citeseer         & 3,327          & 4,732          & 6                & 3,703               \\  
Photo            & 7,487          & 119,043        & 8                & 745                 \\ 
Computers        & 13,752         & 245,861        & 10               & 767               \\  
PubMed           & 19,717         & 44,338         & 3                & 500               \\   \bottomrule 
\end{tabular}  
\end{table}

\begin{table}[t]
  \centering
  \caption{Attack Result}
  \label{AttackResult}
  \fontsize{9pt}{11pt}\selectfont
    \begin{tabular}{cccc}
    \toprule
    \textbf{Dataset} & \textbf{Ptb Rate(\%)} & \textbf{Metattack(\%)} & \textbf{PGD(\%)} \\
    \midrule
    \multirow{5}{*}{\textbf{Cora}}
    & 5 & 79.23±0.50 & 78.22±0.42 \\
    & 10 & 74.28±0.47 & 76.80±0.43 \\
    & 15 & 65.94±0.70 & 74.98±0.23 \\
    & 20 & 55.61±0.92 & 73.50±0.32 \\
    & 25 & 52.58±0.76 & 72.04±0.34 \\

    \midrule
    \multirow{5}{*}{\textbf{Citeseer}}
    & 5 & 63.63±1.28 & 65.82±1.20 \\
    & 10 & 60.13±0.45 & 63.76±1.47 \\
    & 15 & 55.43±0.60 & 61.58±0.83 \\
    & 20 & 51.19±0.86 & 60.58±1.59 \\
    & 25 & 47.94±0.66 & 58.23±1.48 \\

    \midrule
    \multirow{5}{*}{\textbf{Photo}}
    & 5 & 90.83±0.75 & 85.42±0.25 \\
    & 10 & 85.98±0.21 & 81.70±0.65 \\
    & 15 & 75.34±2.97 & 78.92±0.78 \\
    & 20 & 60.44±5.17 & 77.62±0.86 \\
    & 25 & 59.01±3.89 & 75.30±1.96 \\

    \midrule
    \multirow{5}{*}{\textbf{Computers}}
    & 5  & 83.94±0.55 & 79.84±1.60 \\
    & 10 & 75.43±2.07 & 73.01±1.61 \\
    & 15 & 70.53±6.06 & 69.35±2.41 \\
    & 20 & 66.27±6.69 & 65.61±1.86 \\
    & 25 & 67.17±3.74 & 63.54±3.01 \\

    \midrule
    \multirow{5}{*}{\textbf{Pubmed}}
    & 5  & 81.50±0.30 & 82.87±0.15 \\
    & 10 & 68.06±0.55 & 80.16±0.07 \\
    & 15 & 59.65±0.89 & 77.81±0.08 \\
    & 20 & 54.39±1.16 & 75.81±0.07 \\
    & 25 & 51.72±0.45 & 73.75±0.06 \\

    \bottomrule
    \end{tabular}

\end{table}

\begin{table*}[t]
  \centering
  \caption{Node Classification Performance under Metattack(\%)}
  \label{Performance under Metattack}
  \fontsize{10pt}{12pt}\selectfont

  \resizebox{\textwidth}{!}{
    \begin{tabular}{ccc|ccccccc}
    \toprule
    DataSet & Ptb(\%) & GCN & Jaccard & RobustGCN & GNNGuard & Pro-GNN & MidGCN & NoisyGCN & Perseus \\
    \midrule
    \multirow{6}{*}{Cora}
     & 0 & 82.43±0.76 & 80.53±0.73$^6$ & 82.04±0.55$^4$ & 75.63±1.38$^7$ & 81.56±2.01$^5$ & 82.87±0.31$^1$ & 82.64±0.75$^2$ & 82.50±0.66$^3$ \\
     & 5 & 79.23±0.50 & 79.30±0.51$^3$ & 78.31±0.70$^6$ & 73.79±1.38$^7$ & 79.18±1.30$^4$ & 80.54±0.74$^2$ & 78.81±0.50$^5$ & \textbf{81.95±0.41$^1$} \\
     & 10 & 74.28±0.47 & 76.78±0.37$^3$ & 73.07±0.92$^6$ & 73.51±1.34$^5$ & 76.46±2.34$^4$ & 78.78±0.75$^2$ & 72.81±0.59$^7$ & \textbf{79.11±0.41$^1$} \\
     & 15 & 65.94±0.70 & 71.99±0.76$^4$ & 64.91±1.01$^6$ & 69.97±1.89$^5$ & 72.11±1.43$^3$ & 76.09±0.34$^2$ & 64.24±1.09$^7$ & \textbf{76.48±1.06$^1$} \\
     & 20 & 55.61±0.92 & 66.32±0.76$^4$ & 54.60±0.78$^6$ & 66.54±1.91$^3$ & 58.35±1.25$^5$ & 70.44±0.83$^2$ & 55.28±1.70$^7$ & \textbf{73.89±1.59$^1$} \\
     & 25 & 52.58±0.76 & 63.16±0.65$^4$ & 50.70±0.66$^6$ & 64.34±2.62$^2$ & 53.01±0.78$^5$ & 67.88±0.86$^2$ & 52.19±1.20$^7$ & \textbf{70.48±1.07$^1$} \\

    \cline{1-10}
    \multirow{6}{*}{Citeseer} 
     & 0 & 67.62±1.40 & 66.52±0.79$^6$ & 68.91±0.89$^5$ & 65.62±1.04$^7$ & 68.95±0.79$^3$ & \textbf{72.24±0.84$^1$} & 69.11±1.48$^4$ & 71.37±1.16$^2$ \\
     & 5 & 63.63±1.28 & 66.22±1.16$^4$ & 63.81±0.88$^7$ & 65.48±1.12$^5$ & 70.30±0.87$^2$ & \textbf{71.08±1.23$^1$} & 64.80±0.76$^6$ & 70.34±1.25$^3$ \\
     & 10 & 60.13±0.45 & 64.72±0.95$^5$ & 58.66±0.70$^7$ & 65.16±0.93$^4$ & 68.38±0.45$^2$ & 68.42±0.61$^3$ & 59.82±1.95$^6$ & \textbf{69.67±1.16$^1$} \\
     & 15 & 55.43±0.60 & 64.02±0.97$^5$ & 52.17±0.67$^7$ & 65.26±2.09$^4$ & 66.93±0.77$^2$ & 65.65±1.23$^3$ & 54.19±1.12$^6$ & \textbf{69.23±1.05$^1$} \\
     & 20 & 51.19±0.86 & 63.03±1.41$^5$ & 47.55±0.70$^7$ & 63.23±1.78$^4$ & 67.04±0.65$^2$ & 64.73±1.47$^3$ & 50.01±1.29$^6$ & \textbf{67.56±0.90$^1$} \\
     & 25 & 47.94±0.66 & 62.47±1.08$^5$ & 44.46±0.91$^7$ & 64.61±2.16$^3$ & 66.56±0.68$^2$ & 63.44±0.93$^4$ & 45.28±0.79$^6$ & \textbf{67.64±1.53$^1$} \\
    \cline{1-10}

    \multirow{6}{*}{Photo} 
     & 0 & 93.72±0.20 & 93.36±0.18$^2$ & 93.39±0.30$^1$ & 93.04±0.27$^3$ & 83.92±0.31$^6$ & 82.75±0.40$^7$ & 92.51±0.44$^5$ & 92.82±0.18$^4$ \\
     & 5 & 90.83±0.75 & \textbf{92.28±0.35$^1$} & 89.90±0.88$^4$ & 90.30±1.27$^3$ & 81.17±0.36$^5$ & 71.39±0.45$^7$ & 80.20±10.75$^6$ & 91.62±0.41$^2$ \\
     & 10 & 85.98±0.21 & 89.78±0.43$^2$ & 86.00±1.38$^3$ & 83.59±1.84$^4$ & 79.50±0.47$^5$ & 62.72±0.48$^7$ & 66.96±12.85$^6$ & \textbf{90.82±0.31$^1$} \\
     & 15 & 75.34±2.97 & 85.78±1.33$^2$ & 74.53±4.24$^4$ & 71.97±5.43$^5$ & 75.23±0.28$^3$ & 55.60±0.72$^7$ & 63.27±8.11$^6$ & \textbf{90.77±0.53$^1$} \\
     & 20 & 60.44±5.17 & 83.84±1.70$^2$ & 65.55±3.73$^4$ & 62.45±4.95$^5$ & 73.70±0.47$^3$ & 51.03±2.26$^7$ & 53.10±10.19$^6$ & \textbf{89.97±0.46$^1$} \\
     & 25 & 59.01±3.89 & 82.69±1.22$^2$ & 65.60±6.16$^3$ & 54.81±9.97$^5$ & 55.21±1.04$^4$ & 50.45±1.68$^7$ & 54.39±9.10$^6$ & \textbf{88.63±0.27$^1$} \\

    \cline{1-10}
    \multirow{6}{*}{Computers} 
    & 0 & 88.63±0.39 & \textbf{88.85±0.47$^1$} & 85.91±2.19$^4$ & 88.35±0.39$^3$ & * & 68.61±0.21$^6$ & 82.78±3.61$^5$ & 88.47±0.43$^2$ \\
    & 5 & 83.94±0.55 & 83.61±0.82$^2$ & 72.45±2.19$^4$ & 82.18±0.66$^3$ & * & 60.26±0.69$^5$ & 48.24±17.64$^6$ & \textbf{85.17±0.52$^1$} \\
    & 10 & 75.43±2.07 & 75.24±2.49$^3$ & 54.31±9.62$^4$ & 75.63±0.59$^2$ & * & 50.38±0.91$^5$ & 40.82±14.16$^6$ & \textbf{81.86±0.94$^1$} \\
    & 15 & 70.53±6.06 & 71.38±6.28$^3$ & 46.23±6.14$^5$ & 74.21±0.55$^2$ & * & 46.88±0.57$^4$ & 32.12±20.35$^6$ & \textbf{81.59±0.70$^1$} \\
    & 20 & 66.27±6.69 & 67.53±3.37$^3$ & 44.09±4.62$^4$ & 72.91±0.45$^2$ & * & 42.92±0.69$^5$ & 35.55±13.18$^6$ & \textbf{81.17±0.71$^1$} \\
    & 25 & 67.17±3.74 & 67.90±4.75$^3$ & 44.75±7.60$^4$ & 71.95±0.54$^2$ & * & 39.64±0.69$^5$ & 28.10±16.89$^6$ & \textbf{81.24±0.74$^1$} \\

    \cline{1-10}
    \multirow{6}{*}{Pubmed}  
    & 0 & 86.10±0.15 & \textbf{86.06±0.12$^1$} & 85.37±0.15$^3$ & 83.98±0.31$^6$ & * & 84.73±0.15$^5$ & 84.83±0.09$^4$ & 86.01±0.18$^2$  \\
    & 5 & 81.50±0.30 & 81.75±0.28$^4$ & 81.65±0.39$^5$ & 83.81±0.21$^3$ & * & 84.67±0.15$^2$ & 81.19±0.16$^6$ & \textbf{85.59±0.69$^1$}  \\
    & 10 & 68.06±0.55 & 68.42±0.41$^4$ & 67.92±0.43$^6$ & 83.52±0.52$^2$ & * & 82.87±0.40$^3$ & 68.41±0.51$^5$ & \textbf{85.42±0.80$^1$}  \\
    & 15 & 59.65±0.89 & 60.06±0.84$^4$ & 48.54±0.37$^6$ & 82.72±0.38$^2$ & * & 80.17±0.70$^3$ & 54.43±0.50$^5$ & \textbf{85.52±0.22$^1$}  \\
    & 20 & 54.39±1.16 & 54.51±1.26$^4$ & 31.77±0.32$^6$ & 82.01±0.69$^2$ & * & 76.81±0.91$^3$ & 45.60±0.37$^5$ & \textbf{84.81±1.67$^1$}  \\
    & 25 & 51.72±0.45 & 52.01±0.48$^4$ & 22.50±0.51$^6$ & 81.95±0.63$^2$ & * & 69.13±1.80$^3$ & 40.08±0.47$^5$ & \textbf{85.17±0.43$^1$}  \\

    \bottomrule
    \end{tabular}
 }
\end{table*}

\begin{figure}[t]
\centering
\includegraphics[width=\columnwidth]{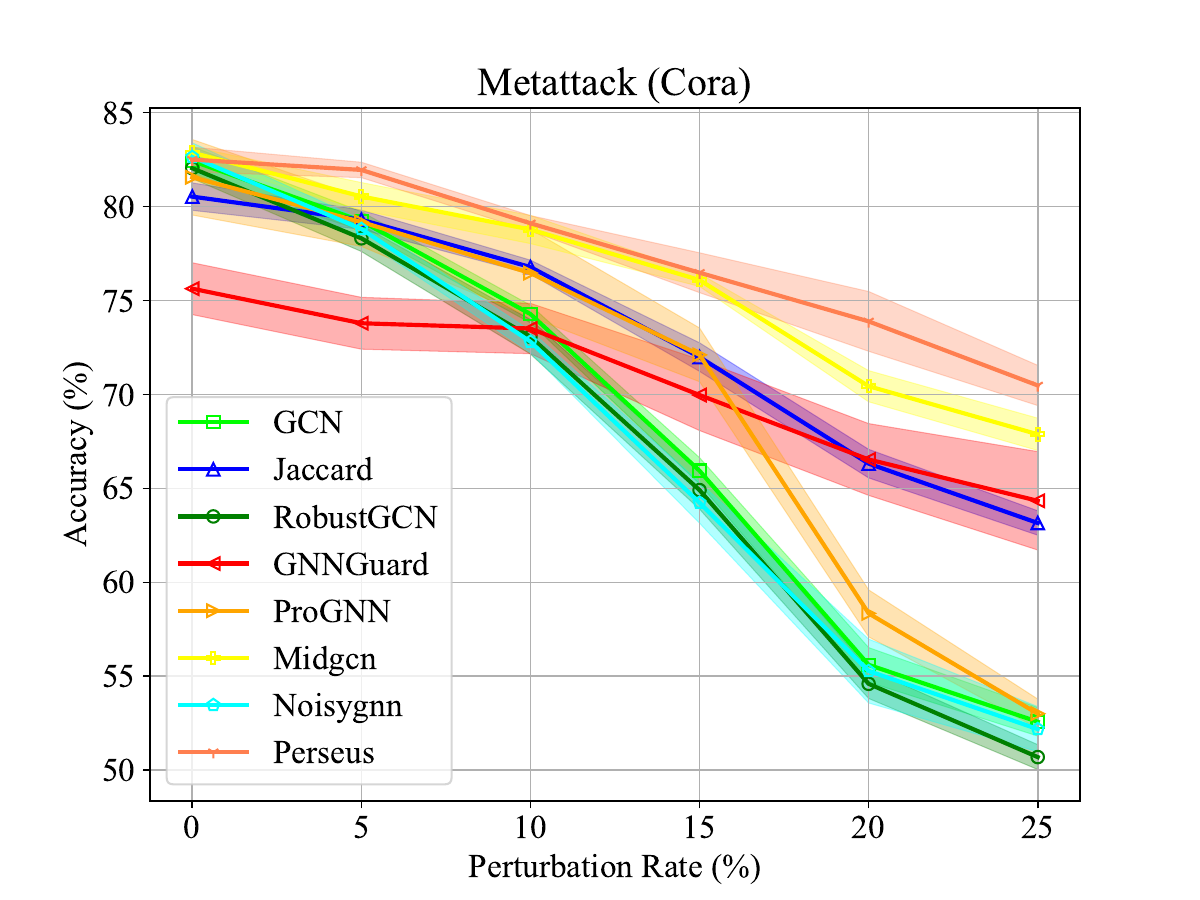} 
\caption{Performance of different models under Metattack with different perturbation rates}
\label{performance}
\end{figure}

\subsubsection{Baseline}

We compared Perseus with seven baseline methods: (1) GCN, (2) GCN-Jaccard, (3) RobustGCN, (4) GNNGuard, (5) Pro-GNN, (6) MidGCN, (7) NoisyGCN. Specifically, GCN-Jaccard \cite{wu2019adversarial} employs Jaccard similarity to reconstruct the graph structure, thereby refining the relationship weights between nodes. RobustGCN \cite{zhou2020robust} combines Gaussian-distributed graph convolutional layers with an attention-based mechanism for information aggregation. GNNGuard \cite{zhang2020gnnguard} enhances the robustness of GNNs through adversarial training and feature regularization. Pro-GNN \cite{jin2020graph} reconstructs the Poisoned Graph to preserve low-rank, sparsity, and the smoothness of attribute features in the graph data. MidGCN \cite{huang2023robust} focuses on the exploitation of mid-frequency signals. NoisyGCN \cite{ennadir2024simple} strengthens the model's robustness against adversarial attacks by introducing random noise into the hidden states of the graph neural network.

\subsection{Experimental Results}
\subsubsection{Performance Comparison} 

We have presented the results of node classification after attacks on five datasets: Cora, Citeseer, Photo, Computers, and Pubmed. The outcomes following the Meta attack are displayed in Table~\ref{Performance under Metattack}, while the results after the PGD attack are shown in Table~\ref{Performance under PGD}, which can be found in Appendix B. Also, the results were visualised, as shown in Figure~\ref{performance}. In light of the findings of the experiment, it can be observed that, firstly, comparing the six state-of-the-art baselines, Perseus essentially outperforms all baseline methods, although it is slightly lower than GCN in clean graphs on the cora dataset. Meanwhile, in graph data with high attack ratio, Perseus exhibits a notable enhancement in comparison to the other baselines. Meanwhile, according on the overall accuracy trend, it can be seen that Perseus can maintain stable performance on graphs with different attack ratios. Its superior performance indicates that the adoption of the curriculum learning strategy, which organises the model's step-by-step learning, is highly beneficial to representation learning.

\begin{table}[t]
  \caption{Performance of Perseus and its variants under Metattack with 0.25 perturbation}
  \label{ablation}
\centering
\fontsize{10pt}{12pt}\selectfont
    \begin{tabular}{cccc}
    \toprule
    \textbf{variants} & \textbf{Cora} & \textbf{Citeseer} & \textbf{Photo} \\ \midrule  
    Perseus             & 70.48±1.07    & 67.64±2.00        & 88.63±0.27     \\ \midrule  
    Cen.              & 57.77±2.98    & 56.44±2.89        & 80.81±1.31     \\  
    Jac.              & 71.73±1.00    & 67.60±0.98        & 90.61±0.51     \\  
    Glo.              & 70.48±1.07    & 67.64±2.00        & 88.63±0.27     \\  

    \bottomrule
    \end{tabular}

\end{table}

\begin{figure}[t]
  \centering
  \includegraphics[width=\columnwidth]{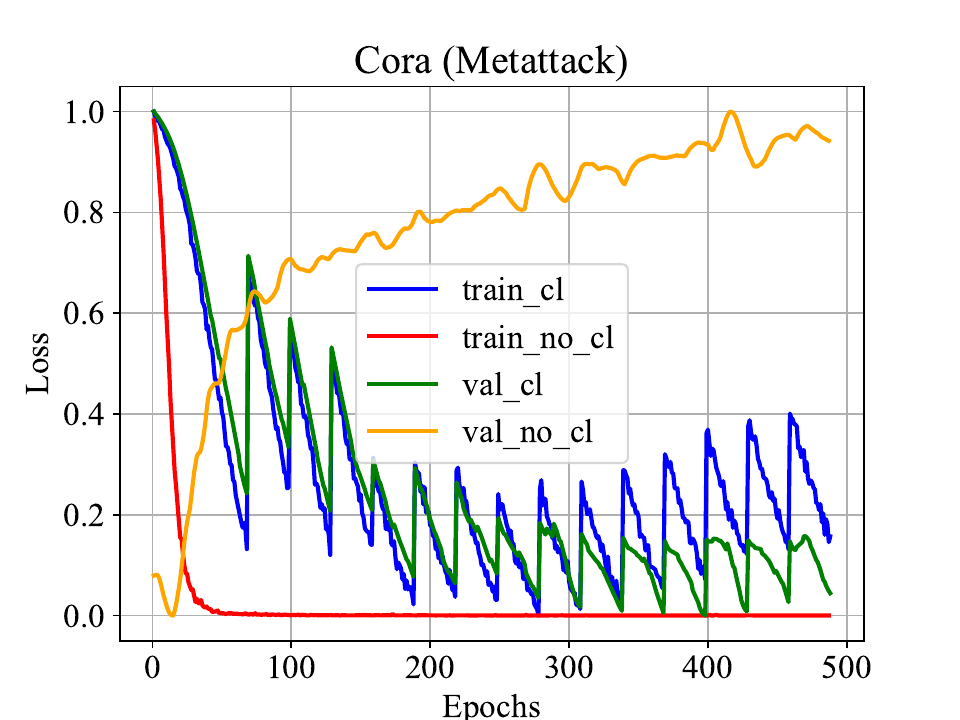} 
  \caption{Train loss and validation loss of Cora under Metattack with 0.25 perturbation.}
  \label{loss_train_val}
\end{figure}

\subsubsection{Ablation Study} 
To examine the contribution of each component and key design in Perseus, we conducted experiments on several variants of Perseus, with the results presented in Table~\ref{ablation}. We initially quantified the edges using PageRank centrality (Cen.), Jaccard (Jac.), and GloHom (Glo.). The results indicate that the impact on the model was minimal when using Jaccard and GloHom metrics, whereas the model's performance was inferior when using PageRank centrality to quantify the edges. Observations from the experiments reveal that the PageRank centrality index fails to effectively distinguish perturbed edges, whereas both Jaccard and GloHom exhibit some discriminative ability, albeit with varying degrees. By adopting a curriculum learning strategy, the model first learns important and uncorrupted edges in the graph. Consequently, in the initial phase, the model focuses on learning common data patterns, establishing a foundation to counteract attack data and ensuring stability when faced with attacks. Subsequently, the model introduces edges of higher complexity incrementally, accompanied by weight decay, which mitigates the impact of attack edges on the model's predictive capabilities. This validates the effectiveness of the curriculum learning concept in enhancing the robustness of the model.

\subsubsection{Sensibility Analysis} 
The model's performance response to the proportion of preheated edges was investigated, with the preheated edges discretized at a scale of 5\% within the interval [0-0.5]. In the Cora dataset, for graphs with a low proportion of perturbations, the variance in the final model's performance was relatively small, with values of 0.81 (Meta0.05ptb), 0.33 (Meta0.10ptb), and 0.73 (Meta0.15ptb). For graphs with a high percentage of attacks, the variance in the final model's performance was 0.58 (Meta0.20ptb) and 0.73 (Meta0.25ptb). The aforementioned data indicates that the model's performance remains stable with respect to the preheating ratio parameter, regardless of the variation in the attack ratio from low to high.

\subsubsection{Robustness Influence} 
In this study, to evaluate the robustness of the Perseus model, we compared the curriculum learning-based Perseus model (denoted as cl) with the graph autoencoder (GAE) model that does not employ curriculum learning (denoted as no\_cl), and plotted their loss function curves on the training and validation sets, as shown in Figure ~\ref{loss_train_val}. Through the analysis of these curves, we observed that the cl model exhibited higher loss on the training set but relatively lower loss on the validation set, indicating a strong robustness of the cl model. Further inspection revealed that as the training progressed and perturbed data were introduced, the training loss of the cl model gradually exceeded the validation loss. This phenomenon suggests that the cl model is capable of generalizing to common data patterns without overfitting to the perturbed data. In other words, in the presence of "perturbed data," halting learning at the appropriate time and introducing new learning content helps guide the model towards the correct solution space.

\bibliographystyle{ACM-Reference-Format}
\bibliography{main}


\begin{thebibliography}{43}


\ifx \showCODEN    \undefined \def \showCODEN     #1{\unskip}     \fi
\ifx \showDOI      \undefined \def \showDOI       #1{#1}\fi
\ifx \showISBNx    \undefined \def \showISBNx     #1{\unskip}     \fi
\ifx \showISBNxiii \undefined \def \showISBNxiii  #1{\unskip}     \fi
\ifx \showISSN     \undefined \def \showISSN      #1{\unskip}     \fi
\ifx \showLCCN     \undefined \def \showLCCN      #1{\unskip}     \fi
\ifx \shownote     \undefined \def \shownote      #1{#1}          \fi
\ifx \showarticletitle \undefined \def \showarticletitle #1{#1}   \fi
\ifx \showURL      \undefined \def \showURL       {\relax}        \fi
\providecommand\bibfield[2]{#2}
\providecommand\bibinfo[2]{#2}
\providecommand\natexlab[1]{#1}
\providecommand\showeprint[2][]{arXiv:#2}

\bibitem[Bengio et~al\mbox{.}(2009)]%
        {bengio09}
\bibfield{author}{\bibinfo{person}{Yoshua Bengio}, \bibinfo{person}{J{\'e}r{\^o}me Louradour}, \bibinfo{person}{Ronan Collobert}, {and} \bibinfo{person}{Jason Weston}.} \bibinfo{year}{2009}\natexlab{}.
\newblock \showarticletitle{Curriculum learning}. In \bibinfo{booktitle}{\emph{Proceedings of the 26th annual international conference on machine learning}}. \bibinfo{pages}{41--48}.
\newblock


\bibitem[Chen et~al\mbox{.}(2019)]%
        {chen2019can}
\bibfield{author}{\bibinfo{person}{Jinyin Chen}, \bibinfo{person}{Yangyang Wu}, \bibinfo{person}{Xiang Lin}, {and} \bibinfo{person}{Qi Xuan}.} \bibinfo{year}{2019}\natexlab{}.
\newblock \showarticletitle{Can adversarial network attack be defended?}
\newblock \bibinfo{journal}{\emph{arXiv preprint arXiv:1903.05994}} (\bibinfo{year}{2019}).
\newblock


\bibitem[Chen et~al\mbox{.}(2020)]%
        {chen2020iterative}
\bibfield{author}{\bibinfo{person}{Yu Chen}, \bibinfo{person}{Lingfei Wu}, {and} \bibinfo{person}{Mohammed Zaki}.} \bibinfo{year}{2020}\natexlab{}.
\newblock \showarticletitle{Iterative deep graph learning for graph neural networks: Better and robust node embeddings}.
\newblock \bibinfo{journal}{\emph{Advances in neural information processing systems}}  \bibinfo{volume}{33} (\bibinfo{year}{2020}), \bibinfo{pages}{19314--19326}.
\newblock


\bibitem[Chu et~al\mbox{.}(2021)]%
        {chu2021cuco}
\bibfield{author}{\bibinfo{person}{Guanyi Chu}, \bibinfo{person}{Xiao Wang}, \bibinfo{person}{Chuan Shi}, {and} \bibinfo{person}{Xunqiang Jiang}.} \bibinfo{year}{2021}\natexlab{}.
\newblock \showarticletitle{CuCo: Graph Representation with Curriculum Contrastive Learning.}. In \bibinfo{booktitle}{\emph{IJCAI}}. \bibinfo{pages}{2300--2306}.
\newblock


\bibitem[Dai et~al\mbox{.}(2018)]%
        {dai2018adversarial}
\bibfield{author}{\bibinfo{person}{Hanjun Dai}, \bibinfo{person}{Hui Li}, \bibinfo{person}{Tian Tian}, \bibinfo{person}{Xin Huang}, \bibinfo{person}{Lin Wang}, \bibinfo{person}{Jun Zhu}, {and} \bibinfo{person}{Le Song}.} \bibinfo{year}{2018}\natexlab{}.
\newblock \showarticletitle{Adversarial attack on graph structured data}. In \bibinfo{booktitle}{\emph{International conference on machine learning}}. PMLR, \bibinfo{pages}{1115--1124}.
\newblock


\bibitem[Deng et~al\mbox{.}(2023)]%
        {deng2023batch}
\bibfield{author}{\bibinfo{person}{Zhijie Deng}, \bibinfo{person}{Yinpeng Dong}, {and} \bibinfo{person}{Jun Zhu}.} \bibinfo{year}{2023}\natexlab{}.
\newblock \showarticletitle{Batch virtual adversarial training for graph convolutional networks}.
\newblock \bibinfo{journal}{\emph{AI Open}}  \bibinfo{volume}{4} (\bibinfo{year}{2023}), \bibinfo{pages}{73--79}.
\newblock


\bibitem[Elman(1993)]%
        {elman1993learning}
\bibfield{author}{\bibinfo{person}{Jeffrey~L Elman}.} \bibinfo{year}{1993}\natexlab{}.
\newblock \showarticletitle{Learning and development in neural networks: The importance of starting small}.
\newblock \bibinfo{journal}{\emph{Cognition}} \bibinfo{volume}{48}, \bibinfo{number}{1} (\bibinfo{year}{1993}), \bibinfo{pages}{71--99}.
\newblock


\bibitem[Ennadir et~al\mbox{.}(2024)]%
        {ennadir2024simple}
\bibfield{author}{\bibinfo{person}{Sofiane Ennadir}, \bibinfo{person}{Yassine Abbahaddou}, \bibinfo{person}{Johannes~F Lutzeyer}, \bibinfo{person}{Michalis Vazirgiannis}, {and} \bibinfo{person}{Henrik Bostr{\"o}m}.} \bibinfo{year}{2024}\natexlab{}.
\newblock \showarticletitle{A Simple and Yet Fairly Effective Defense for Graph Neural Networks}. In \bibinfo{booktitle}{\emph{Proceedings of the AAAI Conference on Artificial Intelligence}}, Vol.~\bibinfo{volume}{38}. \bibinfo{pages}{21063--21071}.
\newblock


\bibitem[Entezari et~al\mbox{.}(2020)]%
        {entezari2020all}
\bibfield{author}{\bibinfo{person}{Negin Entezari}, \bibinfo{person}{Saba~A Al-Sayouri}, \bibinfo{person}{Amirali Darvishzadeh}, {and} \bibinfo{person}{Evangelos~E Papalexakis}.} \bibinfo{year}{2020}\natexlab{}.
\newblock \showarticletitle{All you need is low (rank) defending against adversarial attacks on graphs}. In \bibinfo{booktitle}{\emph{Proceedings of the 13th international conference on web search and data mining}}. \bibinfo{pages}{169--177}.
\newblock


\bibitem[Gong et~al\mbox{.}(2015)]%
        {gong2015curriculum}
\bibfield{author}{\bibinfo{person}{Tieliang Gong}, \bibinfo{person}{Qian Zhao}, \bibinfo{person}{Deyu Meng}, {and} \bibinfo{person}{Zongben Xu}.} \bibinfo{year}{2015}\natexlab{}.
\newblock \showarticletitle{Why curriculum learning \& self-paced learning work in big/noisy data: A theoretical perspective}.
\newblock \bibinfo{journal}{\emph{Big Data \& Information Analytics}} \bibinfo{volume}{1}, \bibinfo{number}{1} (\bibinfo{year}{2015}), \bibinfo{pages}{111--127}.
\newblock


\bibitem[Han et~al\mbox{.}(2018)]%
        {han2018co}
\bibfield{author}{\bibinfo{person}{Bo Han}, \bibinfo{person}{Quanming Yao}, \bibinfo{person}{Xingrui Yu}, \bibinfo{person}{Gang Niu}, \bibinfo{person}{Miao Xu}, \bibinfo{person}{Weihua Hu}, \bibinfo{person}{Ivor Tsang}, {and} \bibinfo{person}{Masashi Sugiyama}.} \bibinfo{year}{2018}\natexlab{}.
\newblock \showarticletitle{Co-teaching: Robust training of deep neural networks with extremely noisy labels}.
\newblock \bibinfo{journal}{\emph{Advances in neural information processing systems}}  \bibinfo{volume}{31} (\bibinfo{year}{2018}).
\newblock


\bibitem[Huang et~al\mbox{.}(2023)]%
        {huang2023robust}
\bibfield{author}{\bibinfo{person}{Jincheng Huang}, \bibinfo{person}{Lun Du}, \bibinfo{person}{Xu Chen}, \bibinfo{person}{Qiang Fu}, \bibinfo{person}{Shi Han}, {and} \bibinfo{person}{Dongmei Zhang}.} \bibinfo{year}{2023}\natexlab{}.
\newblock \showarticletitle{Robust mid-pass filtering graph convolutional networks}. In \bibinfo{booktitle}{\emph{Proceedings of the ACM Web Conference 2023}}. \bibinfo{pages}{328--338}.
\newblock


\bibitem[Ioannidis et~al\mbox{.}(2019)]%
        {ioannidis2019graphsac}
\bibfield{author}{\bibinfo{person}{Vassilis~N Ioannidis}, \bibinfo{person}{Dimitris Berberidis}, {and} \bibinfo{person}{Georgios~B Giannakis}.} \bibinfo{year}{2019}\natexlab{}.
\newblock \showarticletitle{Graphsac: Detecting anomalies in large-scale graphs}.
\newblock \bibinfo{journal}{\emph{arXiv preprint arXiv:1910.09589}} (\bibinfo{year}{2019}).
\newblock


\bibitem[Jiang et~al\mbox{.}(2014)]%
        {jiang14}
\bibfield{author}{\bibinfo{person}{Lu Jiang}, \bibinfo{person}{Deyu Meng}, \bibinfo{person}{Shoou-I Yu}, \bibinfo{person}{Zhenzhong Lan}, \bibinfo{person}{Shiguang Shan}, {and} \bibinfo{person}{Alexander Hauptmann}.} \bibinfo{year}{2014}\natexlab{}.
\newblock \showarticletitle{Self-paced learning with diversity}.
\newblock \bibinfo{journal}{\emph{Advances in neural information processing systems}}  \bibinfo{volume}{27} (\bibinfo{year}{2014}).
\newblock


\bibitem[Jiang et~al\mbox{.}(2015)]%
        {jiang15}
\bibfield{author}{\bibinfo{person}{Lu Jiang}, \bibinfo{person}{Deyu Meng}, \bibinfo{person}{Qian Zhao}, \bibinfo{person}{Shiguang Shan}, {and} \bibinfo{person}{Alexander Hauptmann}.} \bibinfo{year}{2015}\natexlab{}.
\newblock \showarticletitle{Self-paced curriculum learning}. In \bibinfo{booktitle}{\emph{Proceedings of the AAAI Conference on Artificial Intelligence}}, Vol.~\bibinfo{volume}{29}.
\newblock


\bibitem[Jiang et~al\mbox{.}(2018)]%
        {jiang2018mentornet}
\bibfield{author}{\bibinfo{person}{Lu Jiang}, \bibinfo{person}{Zhengyuan Zhou}, \bibinfo{person}{Thomas Leung}, \bibinfo{person}{Li-Jia Li}, {and} \bibinfo{person}{Li Fei-Fei}.} \bibinfo{year}{2018}\natexlab{}.
\newblock \showarticletitle{Mentornet: Learning data-driven curriculum for very deep neural networks on corrupted labels}. In \bibinfo{booktitle}{\emph{International conference on machine learning}}. PMLR, \bibinfo{pages}{2304--2313}.
\newblock


\bibitem[Jin et~al\mbox{.}(2020a)]%
        {jin2020adversarial}
\bibfield{author}{\bibinfo{person}{Wei Jin}, \bibinfo{person}{Yaxin Li}, \bibinfo{person}{Han Xu}, \bibinfo{person}{Yiqi Wang}, {and} \bibinfo{person}{Jiliang Tang}.} \bibinfo{year}{2020}\natexlab{a}.
\newblock \showarticletitle{Adversarial attacks and defenses on graphs: A review and empirical study}.
\newblock \bibinfo{journal}{\emph{arXiv preprint arXiv:2003.00653}} \bibinfo{volume}{10}, \bibinfo{number}{3447556.3447566} (\bibinfo{year}{2020}).
\newblock


\bibitem[Jin et~al\mbox{.}(2020b)]%
        {jin2020graph}
\bibfield{author}{\bibinfo{person}{Wei Jin}, \bibinfo{person}{Yao Ma}, \bibinfo{person}{Xiaorui Liu}, \bibinfo{person}{Xianfeng Tang}, \bibinfo{person}{Suhang Wang}, {and} \bibinfo{person}{Jiliang Tang}.} \bibinfo{year}{2020}\natexlab{b}.
\newblock \showarticletitle{Graph structure learning for robust graph neural networks}. In \bibinfo{booktitle}{\emph{Proceedings of the 26th ACM SIGKDD international conference on knowledge discovery \& data mining}}. \bibinfo{pages}{66--74}.
\newblock


\bibitem[Ju et~al\mbox{.}(2024)]%
        {ju2024comprehensive}
\bibfield{author}{\bibinfo{person}{Wei Ju}, \bibinfo{person}{Zheng Fang}, \bibinfo{person}{Yiyang Gu}, \bibinfo{person}{Zequn Liu}, \bibinfo{person}{Qingqing Long}, \bibinfo{person}{Ziyue Qiao}, \bibinfo{person}{Yifang Qin}, \bibinfo{person}{Jianhao Shen}, \bibinfo{person}{Fang Sun}, \bibinfo{person}{Zhiping Xiao}, {et~al\mbox{.}}} \bibinfo{year}{2024}\natexlab{}.
\newblock \showarticletitle{A comprehensive survey on deep graph representation learning}.
\newblock \bibinfo{journal}{\emph{Neural Networks}} (\bibinfo{year}{2024}), \bibinfo{pages}{106207}.
\newblock


\bibitem[Li et~al\mbox{.}(2023)]%
        {li2023curriculum}
\bibfield{author}{\bibinfo{person}{Haoyang Li}, \bibinfo{person}{Xin Wang}, {and} \bibinfo{person}{Wenwu Zhu}.} \bibinfo{year}{2023}\natexlab{}.
\newblock \showarticletitle{Curriculum graph machine learning: A survey}.
\newblock \bibinfo{journal}{\emph{arXiv preprint arXiv:2302.02926}} (\bibinfo{year}{2023}).
\newblock


\bibitem[Li et~al\mbox{.}(2024)]%
        {li2024graph}
\bibfield{author}{\bibinfo{person}{Xiaohe Li}, \bibinfo{person}{Zide Fan}, \bibinfo{person}{Feilong Huang}, \bibinfo{person}{Xuming Hu}, \bibinfo{person}{Yawen Deng}, \bibinfo{person}{Lei Wang}, {and} \bibinfo{person}{Xinyu Zhao}.} \bibinfo{year}{2024}\natexlab{}.
\newblock \showarticletitle{Graph neural network with curriculum learning for imbalanced node classification}.
\newblock \bibinfo{journal}{\emph{Neurocomputing}}  \bibinfo{volume}{574} (\bibinfo{year}{2024}), \bibinfo{pages}{127229}.
\newblock


\bibitem[Li et~al\mbox{.}(2020)]%
        {li2020deeprobust}
\bibfield{author}{\bibinfo{person}{Yaxin Li}, \bibinfo{person}{Wei Jin}, \bibinfo{person}{Han Xu}, {and} \bibinfo{person}{Jiliang Tang}.} \bibinfo{year}{2020}\natexlab{}.
\newblock \showarticletitle{Deeprobust: A pytorch library for adversarial attacks and defenses}.
\newblock \bibinfo{journal}{\emph{arXiv preprint arXiv:2005.06149}} (\bibinfo{year}{2020}).
\newblock


\bibitem[Luo et~al\mbox{.}(2021)]%
        {luo2021learning}
\bibfield{author}{\bibinfo{person}{Dongsheng Luo}, \bibinfo{person}{Wei Cheng}, \bibinfo{person}{Wenchao Yu}, \bibinfo{person}{Bo Zong}, \bibinfo{person}{Jingchao Ni}, \bibinfo{person}{Haifeng Chen}, {and} \bibinfo{person}{Xiang Zhang}.} \bibinfo{year}{2021}\natexlab{}.
\newblock \showarticletitle{Learning to drop: Robust graph neural network via topological denoising}. In \bibinfo{booktitle}{\emph{Proceedings of the 14th ACM international conference on web search and data mining}}. \bibinfo{pages}{779--787}.
\newblock


\bibitem[Paszke et~al\mbox{.}(2019)]%
        {paszke2019pytorch}
\bibfield{author}{\bibinfo{person}{Adam Paszke}, \bibinfo{person}{Sam Gross}, \bibinfo{person}{Francisco Massa}, \bibinfo{person}{Adam Lerer}, \bibinfo{person}{James Bradbury}, \bibinfo{person}{Gregory Chanan}, \bibinfo{person}{Trevor Killeen}, \bibinfo{person}{Zeming Lin}, \bibinfo{person}{Natalia Gimelshein}, \bibinfo{person}{Luca Antiga}, {et~al\mbox{.}}} \bibinfo{year}{2019}\natexlab{}.
\newblock \showarticletitle{Pytorch: An imperative style, high-performance deep learning library}.
\newblock \bibinfo{journal}{\emph{Advances in neural information processing systems}}  \bibinfo{volume}{32} (\bibinfo{year}{2019}).
\newblock


\bibitem[Said et~al\mbox{.}(2010)]%
        {Said10}
\bibfield{author}{\bibinfo{person}{Alan Said}, \bibinfo{person}{Ernesto De~Luca}, {and} \bibinfo{person}{Sahin Albayrak}.} \bibinfo{year}{2010}\natexlab{}.
\newblock \showarticletitle{How social relationships affect user similarities}.
\newblock \bibinfo{journal}{\emph{Computer Science, Sociology}} (\bibinfo{date}{Jan.} \bibinfo{year}{2010}), \bibinfo{pages}{1--4}.
\newblock


\bibitem[Shrivastava et~al\mbox{.}(2016)]%
        {shrivastava2016training}
\bibfield{author}{\bibinfo{person}{Abhinav Shrivastava}, \bibinfo{person}{Abhinav Gupta}, {and} \bibinfo{person}{Ross Girshick}.} \bibinfo{year}{2016}\natexlab{}.
\newblock \showarticletitle{Training region-based object detectors with online hard example mining}. In \bibinfo{booktitle}{\emph{Proceedings of the IEEE conference on computer vision and pattern recognition}}. \bibinfo{pages}{761--769}.
\newblock


\bibitem[Tang et~al\mbox{.}(2020)]%
        {tang2020transferring}
\bibfield{author}{\bibinfo{person}{Xianfeng Tang}, \bibinfo{person}{Yandong Li}, \bibinfo{person}{Yiwei Sun}, \bibinfo{person}{Huaxiu Yao}, \bibinfo{person}{Prasenjit Mitra}, {and} \bibinfo{person}{Suhang Wang}.} \bibinfo{year}{2020}\natexlab{}.
\newblock \showarticletitle{Transferring robustness for graph neural network against poisoning attacks}. In \bibinfo{booktitle}{\emph{Proceedings of the 13th international conference on web search and data mining}}. \bibinfo{pages}{600--608}.
\newblock


\bibitem[Wei et~al\mbox{.}(2023)]%
        {wei2023clnode}
\bibfield{author}{\bibinfo{person}{Xiaowen Wei}, \bibinfo{person}{Xiuwen Gong}, \bibinfo{person}{Yibing Zhan}, \bibinfo{person}{Bo Du}, \bibinfo{person}{Yong Luo}, {and} \bibinfo{person}{Wenbin Hu}.} \bibinfo{year}{2023}\natexlab{}.
\newblock \showarticletitle{Clnode: Curriculum learning for node classification}. In \bibinfo{booktitle}{\emph{Proceedings of the Sixteenth ACM International Conference on Web Search and Data Mining}}. \bibinfo{pages}{670--678}.
\newblock


\bibitem[Weinshall et~al\mbox{.}(2018)]%
        {weinshall2018curriculum}
\bibfield{author}{\bibinfo{person}{Daphna Weinshall}, \bibinfo{person}{Gad Cohen}, {and} \bibinfo{person}{Dan Amir}.} \bibinfo{year}{2018}\natexlab{}.
\newblock \showarticletitle{Curriculum learning by transfer learning: Theory and experiments with deep networks}. In \bibinfo{booktitle}{\emph{International conference on machine learning}}. PMLR, \bibinfo{pages}{5238--5246}.
\newblock


\bibitem[Wu et~al\mbox{.}(2019)]%
        {wu2019adversarial}
\bibfield{author}{\bibinfo{person}{Huijun Wu}, \bibinfo{person}{Chen Wang}, \bibinfo{person}{Yuriy Tyshetskiy}, \bibinfo{person}{Andrew Docherty}, \bibinfo{person}{Kai Lu}, {and} \bibinfo{person}{Liming Zhu}.} \bibinfo{year}{2019}\natexlab{}.
\newblock \showarticletitle{Adversarial examples on graph data: Deep insights into attack and defense}.
\newblock \bibinfo{journal}{\emph{arXiv preprint arXiv:1903.01610}} (\bibinfo{year}{2019}).
\newblock


\bibitem[Xu et~al\mbox{.}(2019)]%
        {PGD}
\bibfield{author}{\bibinfo{person}{Kaidi Xu}, \bibinfo{person}{Hongge Chen}, \bibinfo{person}{Sijia Liu}, \bibinfo{person}{Pin{-}Yu Chen}, \bibinfo{person}{Tsui{-}Wei Weng}, \bibinfo{person}{Mingyi Hong}, {and} \bibinfo{person}{Xue Lin}.} \bibinfo{year}{2019}\natexlab{}.
\newblock \showarticletitle{Topology Attack and Defense for Graph Neural Networks: An Optimization Perspective}. In \bibinfo{booktitle}{\emph{{IJCAI}}}. \bibinfo{publisher}{ijcai.org}, \bibinfo{pages}{3961--3967}.
\newblock


\bibitem[Xu et~al\mbox{.}(2023)]%
        {xu2023edog}
\bibfield{author}{\bibinfo{person}{Xiaojun Xu}, \bibinfo{person}{Hanzhang Wang}, \bibinfo{person}{Alok Lal}, \bibinfo{person}{Carl~A Gunter}, {and} \bibinfo{person}{Bo Li}.} \bibinfo{year}{2023}\natexlab{}.
\newblock \showarticletitle{Edog: Adversarial edge detection for graph neural networks}. In \bibinfo{booktitle}{\emph{2023 IEEE Conference on Secure and Trustworthy Machine Learning (SaTML)}}. IEEE, \bibinfo{pages}{291--305}.
\newblock


\bibitem[Xu et~al\mbox{.}(2018)]%
        {xu2018characterizing}
\bibfield{author}{\bibinfo{person}{Xiaojun Xu}, \bibinfo{person}{Yue Yu}, \bibinfo{person}{Bo Li}, \bibinfo{person}{Le Song}, \bibinfo{person}{Chengfeng Liu}, {and} \bibinfo{person}{Carl Gunter}.} \bibinfo{year}{2018}\natexlab{}.
\newblock \showarticletitle{Characterizing malicious edges targeting on graph neural networks}.
\newblock  (\bibinfo{year}{2018}).
\newblock


\bibitem[Yang et~al\mbox{.}(2022)]%
        {ICL_SSL}
\bibfield{author}{\bibinfo{person}{Xihong Yang}, \bibinfo{person}{Xiaochang Hu}, \bibinfo{person}{Sihang Zhou}, \bibinfo{person}{Xinwang Liu}, {and} \bibinfo{person}{En Zhu}.} \bibinfo{year}{2022}\natexlab{}.
\newblock \showarticletitle{Interpolation-Based Contrastive Learning for Few-Label Semi-Supervised Learning}.
\newblock \bibinfo{journal}{\emph{IEEE Transactions on Neural Networks and Learning Systems}} (\bibinfo{year}{2022}), \bibinfo{pages}{1--12}.
\newblock
\urldef\tempurl%
\url{https://doi.org/10.1109/TNNLS.2022.3186512}
\showDOI{\tempurl}


\bibitem[Yang et~al\mbox{.}(2024)]%
        {MGCN}
\bibfield{author}{\bibinfo{person}{Xihong Yang}, \bibinfo{person}{Yiqi Wang}, \bibinfo{person}{Yue Liu}, \bibinfo{person}{Yi Wen}, \bibinfo{person}{Lingyuan Meng}, \bibinfo{person}{Sihang Zhou}, \bibinfo{person}{Xinwang Liu}, {and} \bibinfo{person}{En Zhu}.} \bibinfo{year}{2024}\natexlab{}.
\newblock \showarticletitle{Mixed graph contrastive network for semi-supervised node classification}.
\newblock \bibinfo{journal}{\emph{ACM Transactions on Knowledge Discovery from Data}} (\bibinfo{year}{2024}).
\newblock


\bibitem[Zhang and Zitnik(2020)]%
        {zhang2020gnnguard}
\bibfield{author}{\bibinfo{person}{Xiang Zhang} {and} \bibinfo{person}{Marinka Zitnik}.} \bibinfo{year}{2020}\natexlab{}.
\newblock \showarticletitle{Gnnguard: Defending graph neural networks against adversarial attacks}.
\newblock \bibinfo{journal}{\emph{Advances in neural information processing systems}}  \bibinfo{volume}{33} (\bibinfo{year}{2020}), \bibinfo{pages}{9263--9275}.
\newblock


\bibitem[Zhang et~al\mbox{.}(2019)]%
        {zhang2019comparing}
\bibfield{author}{\bibinfo{person}{Yingxue Zhang}, \bibinfo{person}{S Khan}, {and} \bibinfo{person}{Mark Coates}.} \bibinfo{year}{2019}\natexlab{}.
\newblock \showarticletitle{Comparing and detecting adversarial attacks for graph deep learning}. In \bibinfo{booktitle}{\emph{Proc. representation learning on graphs and manifolds workshop, Int. Conf. learning representations, New Orleans, LA, USA}}.
\newblock


\bibitem[Zhao et~al\mbox{.}(2024)]%
        {zhao2024adversarial}
\bibfield{author}{\bibinfo{person}{Kai Zhao}, \bibinfo{person}{Qiyu Kang}, \bibinfo{person}{Yang Song}, \bibinfo{person}{Rui She}, \bibinfo{person}{Sijie Wang}, {and} \bibinfo{person}{Wee~Peng Tay}.} \bibinfo{year}{2024}\natexlab{}.
\newblock \showarticletitle{Adversarial robustness in graph neural networks: A Hamiltonian approach}.
\newblock \bibinfo{journal}{\emph{Advances in Neural Information Processing Systems}}  \bibinfo{volume}{36} (\bibinfo{year}{2024}).
\newblock


\bibitem[Zheng et~al\mbox{.}(2020)]%
        {zheng2020robust}
\bibfield{author}{\bibinfo{person}{Cheng Zheng}, \bibinfo{person}{Bo Zong}, \bibinfo{person}{Wei Cheng}, \bibinfo{person}{Dongjin Song}, \bibinfo{person}{Jingchao Ni}, \bibinfo{person}{Wenchao Yu}, \bibinfo{person}{Haifeng Chen}, {and} \bibinfo{person}{Wei Wang}.} \bibinfo{year}{2020}\natexlab{}.
\newblock \showarticletitle{Robust graph representation learning via neural sparsification}. In \bibinfo{booktitle}{\emph{International Conference on Machine Learning}}. PMLR, \bibinfo{pages}{11458--11468}.
\newblock


\bibitem[Zhou et~al\mbox{.}(2020)]%
        {zhou2020robust}
\bibfield{author}{\bibinfo{person}{Tianyi Zhou}, \bibinfo{person}{Shengjie Wang}, {and} \bibinfo{person}{Jeff Bilmes}.} \bibinfo{year}{2020}\natexlab{}.
\newblock \showarticletitle{Robust curriculum learning: from clean label detection to noisy label self-correction}. In \bibinfo{booktitle}{\emph{International Conference on Learning Representations}}.
\newblock


\bibitem[Zhu et~al\mbox{.}(2019)]%
        {zhu2019robust}
\bibfield{author}{\bibinfo{person}{Dingyuan Zhu}, \bibinfo{person}{Ziwei Zhang}, \bibinfo{person}{Peng Cui}, {and} \bibinfo{person}{Wenwu Zhu}.} \bibinfo{year}{2019}\natexlab{}.
\newblock \showarticletitle{Robust graph convolutional networks against adversarial attacks}. In \bibinfo{booktitle}{\emph{Proceedings of the 25th ACM SIGKDD international conference on knowledge discovery \& data mining}}. \bibinfo{pages}{1399--1407}.
\newblock


\bibitem[Z{\"u}gner et~al\mbox{.}(2018)]%
        {zugner2018adversarial}
\bibfield{author}{\bibinfo{person}{Daniel Z{\"u}gner}, \bibinfo{person}{Amir Akbarnejad}, {and} \bibinfo{person}{Stephan G{\"u}nnemann}.} \bibinfo{year}{2018}\natexlab{}.
\newblock \showarticletitle{Adversarial attacks on neural networks for graph data}. In \bibinfo{booktitle}{\emph{Proceedings of the 24th ACM SIGKDD international conference on knowledge discovery \& data mining}}. \bibinfo{pages}{2847--2856}.
\newblock


\bibitem[Z{\"{u}}gner and G{\"{u}}nnemann(2019)]%
        {Metattack}
\bibfield{author}{\bibinfo{person}{Daniel Z{\"{u}}gner} {and} \bibinfo{person}{Stephan G{\"{u}}nnemann}.} \bibinfo{year}{2019}\natexlab{}.
\newblock \showarticletitle{Adversarial Attacks on Graph Neural Networks via Meta Learning}. In \bibinfo{booktitle}{\emph{{ICLR} (Poster)}}. \bibinfo{publisher}{OpenReview.net}.
\newblock


\end{thebibliography}
\appendix

\section{Mathematical Proof}
We model the global homophily index to identify and eliminate edges that propagate the most misleading information. That is, we aim to find an edge whose removal would maximize $\Delta \mathrm{Hom}$. This can be mathematically formulated as:
\begin{equation}  
\label{equ:4}
\underset{\Delta \mathbf{A}}{\mathrm{argmax}} \Delta \mathrm{Hom}
\end{equation}

Suppose the edge $ e_{kl} $ is removed, resulting in a new adjacency matrix $\mathbf{A}^{\prime}$. Here, $\Delta \mathbf{A} = \mathbf{A}^{\prime} - \mathbf{A}$. For $\Delta \mathbf{A}$, $\Delta \mathbf{A}_{kl} = \Delta \mathbf{A}_{lk} = -1$, with all other values being 0. We further derived and optimized this result to achieve the final outcome.

\begin{equation}  
\label{equ:5}
\Delta \mathrm{Hom} = <M^{\prime} - M, \mathcal{J}> = \alpha <M^{\prime} \Delta \mathbf{A} M, \mathcal{J}>  
\end{equation}

Where $M^{\prime} = (I - \alpha\mathbf{A}^{\prime})^{-1}$. Given this modification, we can determine the values at each position in the matrix $M^{\prime}\Delta \mathbf{A} M$. The calculation process is detailed as follows:

\begin{equation}  
\label{equ:6}
{\left[M^{\prime}\Delta \mathbf{A} M\right]_{i j}} = \sum_{s=1}^{n} \sum_{t=1}^{n} M_{i s}^{\prime} \Delta A_{s t} M_{t j} = -M_{i l}^{\prime} M_{k j} - M_{i k}^{\prime} M_{l j}  
\end{equation}

By substituting Equation (6) into Equation (5), we can perform detailed mathematical derivations and obtain the following results\begin{equation}  
\label{equ:7}
\begin{aligned}  
\Delta \mathrm{Hom} &= \alpha < M^{\prime}\Delta \mathbf{A} M , \mathcal{J}> = \alpha \sum_{i, j}\left[M^{\prime}\Delta \mathbf{A} M\right]_{i j} \mathcal{J}_{i j} \\  
&= -\alpha \sum_{i, j}(M_{i l}^{\prime} M_{k j}+M_{i k}^{\prime} M_{l j})\mathcal{J}_{i j} \\  
&= -\alpha \sum_{i, j}(M_{l i}^{\prime} \mathcal{J}_{i j} M_{j k}+M_{k i}^{\prime} \mathcal{J}_{i j} M_{j l}) \\  
&= -2 \alpha\left[M^{\prime} \mathcal{J} M\right]_{k l}  
\end{aligned}  
\end{equation}

Further derivation leads to a conclusion.\begin{equation}  
\label{equ:8}
\underset{\Delta \mathbf{A}}{\mathrm{argmax}} \Delta  \mathrm{Hom} = \underset{\Delta \mathbf{A}}{\mathrm{argmin}}<M^{\prime} \mathcal{J} M,-\Delta \mathbf{A}>
\end{equation}

\section{Attack Result}
\begin{table*}[t]
  \centering
  \caption{Node Classification Performance under PGD(\%)}
  \label{Performance under PGD}
  \fontsize{10pt}{12pt}\selectfont
  \resizebox{\textwidth}{!}{
    \begin{tabular}{ccc|ccccccc}
    \toprule
    DataSet & Ptb(\%) & GCN & Jaccard & RobustGCN & GNNGuard & Pro-GNN & MidGCN & NoisyGCN & Perseus \\
    \midrule
    \multirow{6}{*}{Cora}
     & 0 & 82.69±0.63 & 80.54±0.57$^6$ & 81.91±0.43$^4$ & 75.63±0.89$^7$ & 81.16±2.17$^5$ & \textbf{82.90±0.27$^1$} & 82.75±0.41$^2$ & 82.28±0.80$^3$ \\
     & 5 & 78.22±0.42 & 77.81±0.47$^5$ & 77.77±0.52$^6$ & 74.96±1.12$^7$ & 78.52±1.38$^3$ & 78.83±0.30$^2$ & 77.99±0.67$^4$ & \textbf{79.05±0.30$^1$} \\
     & 10 & 76.80±0.43 & 76.40±0.64$^3$ & 76.22±0.45$^4$ & 72.69±2.13$^7$ & 75.57±0.48$^6$ & 76.56±0.30$^2$ & 75.63±0.42$^5$ & \textbf{78.49±0.42$^1$} \\
     & 15 & 74.98±0.23 & 74.75±0.33$^3$ & 74.60±0.81$^5$ & 74.71±1.04$^4$ & 74.01±0.58$^7$ & 75.10±0.34$^2$ & 74.35±0.49$^6$ & \textbf{77.53±0.55$^1$} \\
     & 20 & 73.50±0.32 & 73.26±0.62$^3$ & 73.10±0.42$^4$ & 71.99±1.09$7$ & 72.39±0.47$^6$ & 73.40±0.27$^2$ & 72.81±0.33$^5$ & \textbf{75.20±0.43$^1$} \\
     & 25 & 72.04±0.34 & 72.35±0.66$^2$ & 71.91±0.28$^4$ & 70.87±1.02$^7$ & 71.79±0.11$^5$ & 72.20±0.23$^3$ & 71.77±0.41$^6$ & \textbf{74.50±0.53$^1$} \\
    \cline{1-10}

    \multirow{6}{*}{Citeseer}
     & 0 & 68.27±0.95 & 67.65±1.32$^6$ & 69.28±0.72$^5$ & 65.80±1.89$^7$ & 69.43±0.22$^3$ & \textbf{71.97±0.92$^1$} & 69.36±1.16$^4$ & 70.93±1.68$^2$ \\
     & 5 & 65.82±1.20 & 66.27±0.96$^6$ & 66.98±0.37$^5$ & 66.06±1.29$^7$ & 69.51±0.32$^3$ & \textbf{71.02±0.84$^1$} & 67.67±1.34$^4$ & 69.95±1.55$^2$ \\
     & 10 & 63.76±1.47 & 66.27±0.88$^4$ & 64.12±0.71$^7$ & 65.99±0.86$^5$ & 68.22±0.68$^2$ & \textbf{68.52±1.84$^1$} & 64.86±1.57$^6$ & 68.19±1.27$^3$ \\
     & 15 & 61.58±0.83 & 65.18±1.40$^5$ & 62.52±0.95$^7$ & 65.44±1.28$^4$ & 67.16±0.33$^3$ & 67.86±1.25$^2$ & 63.98±1.48$^6$ & \textbf{68.66±0.43$^1$} \\
     & 20 & 60.58±1.59 & 64.18±1.06$^5$ & 61.60±0.32$^7$ & 64.79±0.55$^4$ & 66.96±0.63$^2$ & 66.65±1.04$^3$ & 62.84±1.56$^6$ & \textbf{68.14±1.03$^1$} \\
     & 25 & 58.23±1.48 & 64.45±0.51$^4$ & 60.09±0.90$^7$ & 65.18±1.51$^2$ & 63.27±0.26$^5$ & 64.79±0.81$^3$ & 60.27±1.29$^6$ & \textbf{68.27±1.42$^1$} \\
    \cline{1-10}

    \multirow{6}{*}{Photo} 
     & 0 & 93.71±0.25 & 93.10±0.15$^2$ & \textbf{93.67±0.29$^1$} & 92.97±0.47$^3$ & 83.80±0.40$^6$ & 82.73±0.19$^7$ & 92.47±0.36$^5$ & 92.93±0.10$^4$ \\
     & 5 & 85.42±0.25 & 87.00±0.23$^2$ & 86.09±0.15$^3$ & 85.57±0.38$^4$ & 78.95±0.26$^6$ & 78.19±0.32$^7$ & 84.98±0.34$^5$ & \textbf{87.56±0.61$^1$} \\
     & 10 & 81.70±0.65 & 83.19±0.22$^3$ & 83.45±0.28$^2$ & 83.02±0.26$^4$ & 76.86±0.26$^6$ & 76.13±0.20$^7$ & 82.26±0.56$^5$ & \textbf{84.90±0.62$^1$} \\
     & 15 & 78.92±0.78 & 80.32±0.53$^3$ & 81.60±0.19$^2$ & 80.87±0.37$^4$ & 75.23±0.28$^6$ & 74.83±0.32$^7$ & 80.25±0.76$^5$ & \textbf{83.84±0.43$^1$} \\
     & 20 & 77.62±0.86 & 78.30±1.03$^5$ & 80.08±0.31$^2$ & 79.56±0.41$^3$ & 73.70±0.47$^7$ & 74.42±0.32$^6$ & 78.80±0.54$^4$ & \textbf{81.86±0.57$^1$} \\
     & 25 & 75.30±1.96 & 76.70±0.91$^5$ & 79.14±0.28$^2$ & 78.08±0.47$^3$ & 71.59±0.25$^7$ & 73.65±1.06$^6$ & 77.35±0.89$^4$ & \textbf{81.07±0.99$^1$} \\
    \cline{1-10}

    \multirow{6}{*}{Computers}  
    & 0 & 89.12±0.37 & \textbf{88.71±0.51$^1$} & 85.52±2.03$^4$ & 88.34±0.35$^2$ & * & 68.65±0.31$^6$ & 83.74±2.50$^5$ & 88.32±0.54$^3$ \\
    & 5 & 79.84±1.60 & 79.91±1.12$^3$ & 77.63±2.72$^4$ & 80.41±0.28$^2$ & * & 66.62±0.39$^6$ & 74.42±3.37$^5$ & \textbf{82.54±0.50$^1$} \\
    & 10 & 73.01±1.61 & 73.07±2.15$^4$ & 74.42±1.38$^3$ & 75.02±0.49$^2$ & * & 65.09±0.50$^6$ & 71.42±4.24$^5$ & \textbf{79.74±0.62$^1$} \\
    & 15 & 69.35±2.41 & 67.89±2.470$^4$ & 70.23±2.76$^3$ & 71.45±0.47$^2$ & * & 63.97±0.50$^5$ & 61.87±16.60$^6$ & \textbf{77.35±0.53$^1$} \\
    & 20 & 65.61±1.86 & 65.21±1.53$^4$ & 67.52±3.50$^3$ & 69.79±0.54$^2$ & * & 63.09±1.01$^6$ & 63.81±9.40$^5$ & \textbf{74.81±0.42$^1$} \\
    & 25 & 63.54±3.01 & 64.62±3.34$^4$ & 65.58±2.95$^3$ & 67.86±0.61$^2$ & * & 62.62±0.96$^5$ & 62.09±4.38$^6$ & \textbf{72.44±0.82$^1$} \\

    \cline{1-10}
    \multirow{6}{*}{Pubmed}  
    & 0 & 86.09±0.16 & \textbf{86.07±0.17$^1$} & 85.38±0.15$^3$ & 84.27±0.27$^6$ & * & 84.76±0.10$^5$ & 84.82±0.11$^4$ & 85.60±0.95$^2$  \\
    & 5 & 82.87±0.15 & 82.91±0.16$^3$ & 82.61±0.13$^4$ & 83.54±0.30$^2$ & * & 82.26±0.18$^5$ & 82.08±0.10$^6$ & \textbf{84.52±1.96$^1$}  \\
    & 10 & 80.16±0.07 & 80.24±0.08$^3$ & 80.09±0.07$^4$ & 83.36±0.26$^2$ & * & 80.07±0.15$^5$ & 79.56±0.12$^6$ & \textbf{84.90±0.18$^1$}  \\
    & 15 & 77.81±0.08 & 77.89±0.08$^4$ & 77.76±0.09$^5$ & 82.46±0.17$^2$ & * & 78.50±0.10$^3$ & 77.27±0.07$^6$ & \textbf{84.17±1.25$^1$}  \\
    & 20 & 75.81±0.07 & 75.84±0.06$^4$ & 75.84±0.10$^5$ & 82.02±0.35$^2$ & * & 76.99±0.06$^3$ & 75.25±0.11$^6$ & \textbf{84.01±2.46$^1$}  \\
    & 25 & 73.75±0.06 & 73.81±0.05$^5$ & 73.84±0.15$^4$ & 81.65±0.29$^2$ & * & 75.64±0.12$^3$ & 73.31±0.06$^6$ & \textbf{84.28±0.22$^1$}  \\

    \bottomrule
    \end{tabular}
  }

\end{table*}

\end{document}